# Hivemind: A Federated Protocol for AI-Mediated Tacit Knowledge Sharing in Organisations

Srinivasan Sundaresan
Rakuten Group, Inc.
srinivasan.sundaresan95@gmail.com

## Abstract

This paper presents Hivemind, a federated protocol that enables AI agents to exchange tacit knowledge on behalf of their users, asynchronously and without requiring any direct human interaction. Individual AI memory systems capture the decisions, reasoning, and contextual knowledge that surface during AI-assisted work. Hivemind creates the shared space in which this captured knowledge can flow between individuals across team and departmental boundaries, addressing a structural gap that neither centralised knowledge repositories nor existing AI tooling closes, even where partial solutions such as expert-finding systems exist. The paper evaluates centralised versus decentralised design approaches, proposes a federated architecture grounded in Nonaka and Takeuchi's (1995) Socialisation, Externalisation, Combination and Internalisation (SECI) model, and demonstrates a proof-of-concept implementation using Design Science Research (DSR) methodology. Measuring the enterprise benefit of federated tacit expertise sharing remains an open task for future work.

## 1. Introduction

### *1.1 AI and the Transformation of Workplace Knowledge Management*

Research on AI-assisted knowledge management (KM) is more than two decades old, from expert systems and organisational memory to ontology-based systems (Linger & Burstein, 1998; Abecker et al., 2000), and KM is linked to organisational innovativeness (Chen et al., 2010). Practical deployment at institutional scale grew quickly only from 2022-2023 onwards, driven by the emergence of large language models (LLMs) with sufficient capability for real-world organisational use (McKinsey & Company, 2023). AI-powered assistants now query corporate knowledge management systems (KMS) - wikis (Confluence, GitHub Wikis), version-controlled codebases, project tracking tools, communication platforms, and analytics systems - through standardised integration protocols such as the Model Context Protocol (MCP; Anthropic, 2024). LLMs have demonstrated strong capability in synthesising knowledge across heterogeneous sources (Lewis et al., 2020; Brown et al., 2020), enabling AI agents to combine information from multiple explicit repositories in response to a single query.

This progress corresponds to Nonaka and Takeuchi's (1995) Combination quadrant: AI is now highly effective at recombining established, codified, and retrievable knowledge. Existing AI memory tools also partly cover Externalisation, by capturing the traces of reasoning that surface during AI-assisted work, and Internalisation, by returning an individual's own past context to them, so explicit knowledge handling is comparatively mature within one person's own workspace. What remains largely unaddressed is Socialisation between people, which this paper approaches by letting their AI agents exchange knowledge on their behalf.

### *1.2 The Persistent Gap: Tacit Knowledge*

Yet a substantial portion of workplace knowledge resists this progress. Polanyi's (1966) foundational insight - 'we can know more than we can tell' - describes knowledge that is personal, contextual, and experience-based: the rationale behind architectural decisions, awareness of past failures, team-specific conventions, and informal heuristics that are never written down. This is tacit knowledge, and it constitutes a significant share of what differentiates high-performing teams and individuals (Leonard & Sensiper, 1998; DeLong, 2004).

The problem is not primarily one of deliberate concealment, though that does occur. The majority of tacit knowledge remains unshared for structural and behavioural reasons that are not volitional. Employees do not know who might benefit from their knowledge (Kotlarsky et al., 2015), and there is no institutional mechanism prompting them to share (Burstein et al., 2010). They are also too busy to pause and articulate what they know (Foos et al., 2006), or work within social dynamics - particularly in technical or remote-first workplaces - that do not facilitate the informal exchanges which historically enable tacit knowledge transfer (Hansen, 1999; Nonaka & Von Krogh, 2009). In IT and software organisations specifically, employees tend towards lower rates of spontaneous informal interaction compared to co-located, physical workplaces (Valk & Planojevic, 2021; Abidi et al., 2023), compounding these barriers.

Where deliberate withholding does occur, it is well-documented and distinct. Connelly et al. (2012) defined and measured knowledge hiding - deliberate concealment of knowledge from colleagues while still employed - finding it widespread. Skerlavaj et al.'s (2023) meta-analysis confirmed this remains a persistent organisational challenge. Anand et al. (2020) identified drivers of knowledge hiding including competition, psychological ownership, and hostility or abuse.

The consequences are tangible regardless of cause: parallel rediscovery of known solutions, repeated mistakes, extended onboarding periods, and fragile dependency on specific individuals (DeLong, 2004; Figure 1). Burstein et al. (2010) demonstrated the gap between KM investment and actual knowledge-sharing outcomes, finding KM roles poorly defined in practice. Owen and Burstein (2005) noted that tacit knowledge is embedded in project work and rarely surfaces to formal KMS.

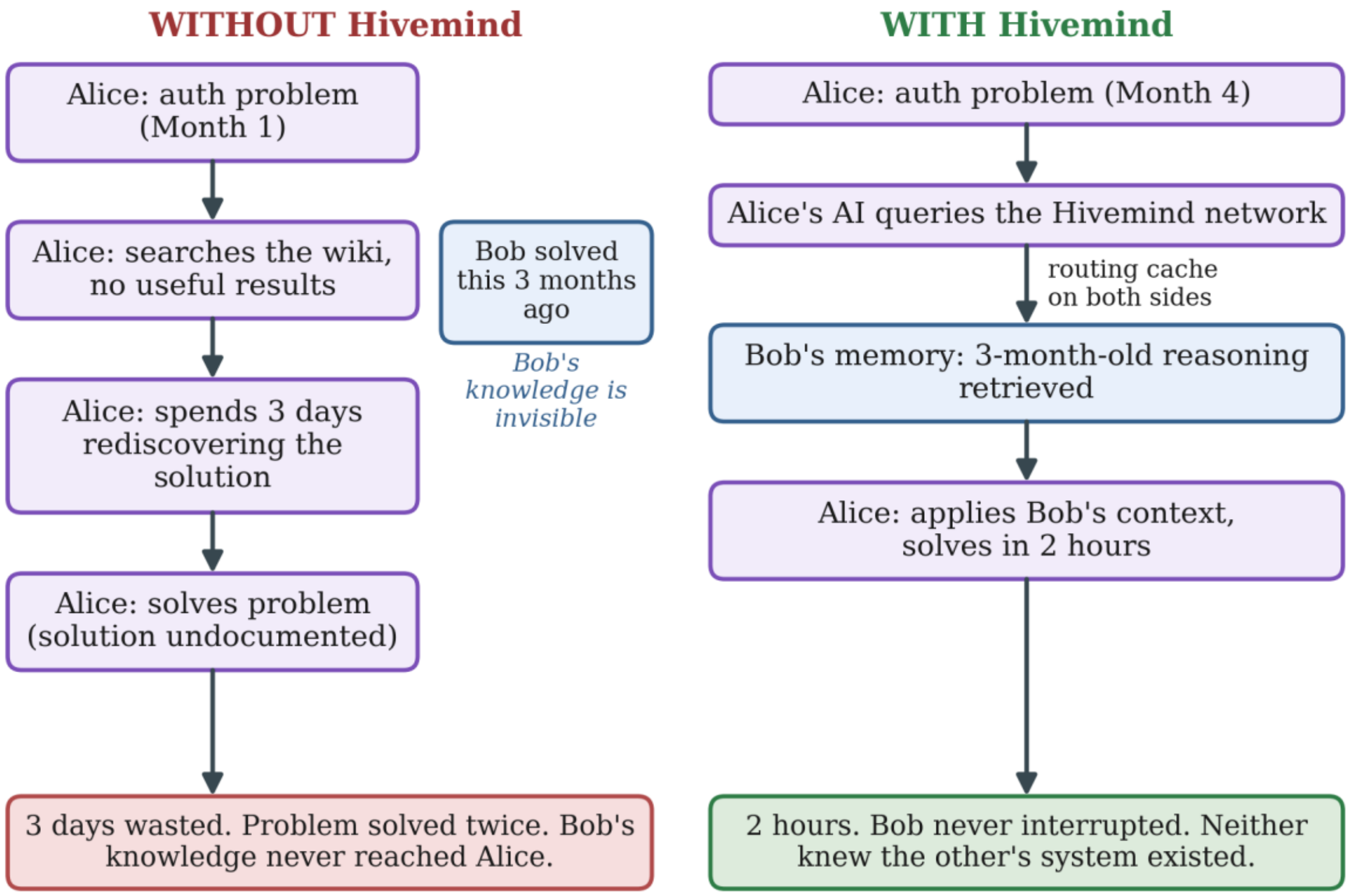

*Figure 1: The parallel rediscovery problem (left) and its resolution via Hivemind (right). Without Hivemind, equivalent problems are solved independently; with Hivemind, existing knowledge is surfaced asynchronously and attributed.*

### *1.3 Why Socialisation Fails in Digital Organisations*

Nonaka and Takeuchi's (1995) Socialisation, Externalisation, Combination and Internalisation (SECI) model places Socialisation - the tacit-to-tacit transfer of knowledge through shared experience and co-presence - at the heart of how organisations create knowledge. It is also the mode that digital organisations consistently fail to support.

Hansen (1999), in a foundational study of inter-unit knowledge sharing, identified the search-transfer problem: weak ties aid discovery of who holds relevant knowledge but impede the actual transfer of complex tacit knowledge. Johannessen et al. (2001) argued that IT is fundamentally inadequate for tacit knowledge management and may actively worsen it. Stenmark (2000) argued that attempts to make tacit knowledge explicit often fail, and showed that recommender technology can nonetheless help make tacit knowledge available to the organisation.

Behavioural factors compound the problem. High workloads leave little time for informal knowledge exchange (Foos et al., 2006). Social reticence and limited informal interaction - particularly common in engineering-focused workplaces and remote-first environments (Valk & Planojevic, 2021; Abidi et al., 2023) - further reduce the spontaneous socialisation that has historically been the primary vehicle for tacit knowledge transfer (Nonaka & Von Krogh, 2009). Williams (2012), in a case study of two integrated health and social care services, found that trust, interpersonal relationships and co-location of facilities were important for tacit knowledge exchange.

A further systemic barrier, particularly relevant in multinational and global organisations, is linguistic diversity. Even when knowledge exists across teams, language differences impede both its articulation and its discovery. AI-mediated knowledge exchange could additionally support automatic translation, enabling tacit-knowledge residues captured in one language to be surfaced and interpreted in another without human translation effort.

### *1.4 Transactive Memory Systems and Their Limitations*

One partial solution to expertise unawareness is the Transactive Memory System (TMS), introduced by Wegner (1987) and situated within the broader organisational memory literature (Stein, 1995): groups develop shared awareness of who knows what, enabling efficient query routing without requiring all members to know everything. Lewis et al. (2005) and Kotlarsky et al. (2015) extended TMS theory to organisational teams.

In practice, TMS has a straightforward problem: it requires people to keep it updated. As workloads increase, maintenance stops. The directory grows stale, and the system is quietly abandoned (Peltokorpi, 2008). The maintenance burden is not incidental; it is the primary reason TMS has not scaled beyond small, stable teams.

### *1.5 The Opportunity: AI Agents in Ba*

AI memory systems represent a qualitatively new development, part of a broader shift towards agentic AI systems operating with increasing autonomy across domains (Li et al., 2025). Such systems automatically capture tacit knowledge - decisions made, context behind choices, architectural reasoning, lessons from past failures - as AI-observable episodic memory, without requiring deliberate human curation. More precisely, AI episodic memory captures tacit-knowledge residues - the partially externalised traces of contextual experience that surface as individuals work with an AI system - rather than the full Polanyian tacit dimension. Collins (2010) distinguishes at least three forms of tacit knowledge - relational, somatic, and collective - and only relational tacit knowledge, the contextual and interactional knowing embedded in individual practice and exchange, falls within the scope of what episodic AI memory can begin to approximate; somatic embodied skill and collectively-held enterprise knowledge remain inherently beyond its reach.

Consider a concrete example. Alice is working through an authentication integration problem and asks her AI assistant for relevant context. Her assistant queries Bob's knowledge node via Hivemind - surfacing reasoning Bob recorded three months ago when solving the same problem - and returns it to Alice attributed to its source. Alice has never spoken to Bob about this. Bob is not interrupted. No central system stored the knowledge. The tacit context that would otherwise have required a scheduled knowledge-transfer meeting, or would simply never have been found, is retrieved asynchronously, on demand, by the AI, as shown schematically in Figure 2.

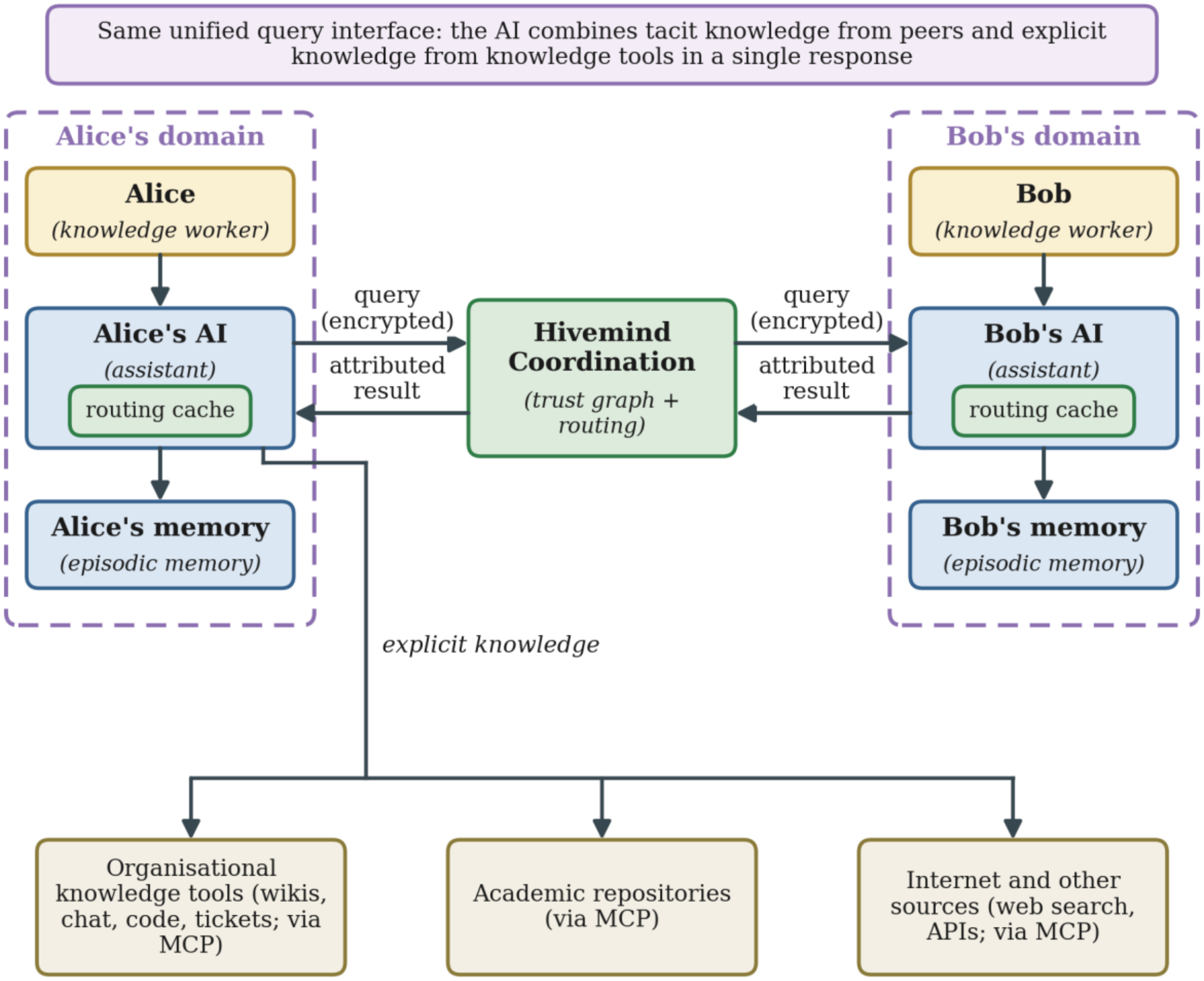

*Figure 2: The Hivemind knowledge ecosystem - AI agents query tacit knowledge from peers and explicit knowledge from workplace tools through a unified interface.*

He et al. (2026) proposed, in a conceptual analysis, that generative AI transforms knowledge creation at the individual level - examining the dyadic relationship between a human and their AI assistant within a personal Ba (the shared context in which knowledge is created; Nonaka & Konno, 1998). This paper extends that frame to a fundamentally different scope: multi-agent, multi-individual Ba, in which multiple AI agents act on behalf of multiple human users across a federated knowledge space, enabling asynchronous tacit knowledge exchange between individuals who may never directly interact. It proposes Hivemind, a federated protocol that implements this vision.

## 2. Background and Related Work

### *2.1 Explicit and Tacit Knowledge*

The distinction between tacit and explicit knowledge comes from Polanyi (1966). His point, bluntly stated: we know more than we can tell. Explicit knowledge - the kind you can write down - is transferable and codifiable. Tacit knowledge is personal, contextual, accumulated through experience, and resistant to articulation. Nonaka and Takeuchi (1995) built the SECI model on this foundation, describing four conversion modes - Socialisation (tacit-to-tacit), Externalisation (tacit-to-explicit), Combination (explicit-to-explicit), and Internalisation (explicit-to-tacit) - as the engine of knowledge creation in organisations.

Nonaka and Konno (1998) introduced Ba: the shared context in which knowledge is created. The key insight is that knowledge cannot be cleanly separated from the setting in which it emerged - strip out the Ba and something essential is lost. Different Ba types correspond to different SECI phases.

Nonaka et al. (2000) defined four types of Ba, each corresponding to a SECI phase: Originating Ba (face-to-face, individual - the space of Socialisation), Dialoguing Ba (collective, face-to-face - Externalisation), Systemising Ba (virtual, collective - Combination), and Exercising Ba (virtual, individual - Internalisation). Digital platforms have long supported Systemising Ba, enabling virtual recombination of explicit knowledge. They have largely failed at Originating Ba, the co-present experience-sharing through which tacit knowledge actually moves between people. Faraj et al. (2011) note that meaningful knowledge exchange in virtual communities depends on boundary-spanning participation and temporal flexibility - conditions that are inherently difficult to engineer. Hivemind's contribution is enabling the tacit-knowledge-sharing dynamic of Originating Ba through Systemising Ba infrastructure: the co-presence requirement is removed because AI agents act as proxies, asynchronously. This cross-Ba-type flow, mapped in Section 7.1, is not envisaged in Nonaka et al. (2000) and represents, to the author's knowledge, the paper's principal theoretical extension.

### *2.2 Cross-Nomenclature as a Structural Barrier*

There is a quieter barrier that is easy to overlook: teams that work in isolation long enough, particularly in larger organisations where units have fewer incidental contacts, develop their own vocabulary. Different units coin different names for the same concepts - a natural outcome of forming separate communities of practice (Wenger, 1998) - and when those teams eventually try to share knowledge, the terminology mismatch silently breaks retrieval. Edgington et al. (2004) proposed controlled vocabulary as a fix. Hameed et al. (2004) documented systematic ontology mismatches between business units. Blomqvist and Öhgren (2008) found that separate units simply 'do not use the same terminology,' meaning queries fail even when the relevant knowledge exists somewhere in the network. Any knowledge-sharing system that relies on exact name matching will hit this wall.

### *2.3 Prior Work in Federated and Peer-to-Peer Knowledge Systems*

Bonifacio et al. (2002, 2004) proposed foundational peer-to-peer KM architectures, demonstrating that local autonomy enables richer knowledge capture than centralised systems. Maier and Hädrich (2006) directly compared centralised and peer-to-peer KMS, finding peer-to-peer systems superior in resilience and scalability but noting higher search complexity. Zhuge (2002) proposed a knowledge flow model for peer-to-peer team knowledge sharing and management.

In the multi-agent AI space, MELD (Lovén et al., 2026) is among the most closely related prior work - a protocol for merging knowledge across distributed agentic memories. MELD reports empirical results, but it synchronises agents' knowledge-graph memories under a shared group key, with no pairwise trust handshake, per-person capability tokens, blind relay, alias bridging or referral routing. He et al. (2026) explored AI and Ba at the individual human-AI dyad level, which is a useful conceptual starting point but a different scope from what is proposed here. Other recent work addresses neighbouring problems. Multi-Agent Transactive Memory stores trajectories produced by many agents in a shared repository for reuse by other agents (Kim et al., 2026), and a shared organisational memory, deployed in selected teams at one company and not yet evaluated for usefulness, captures experience from enterprise coding agents for reuse by future agents (Dhanyamraju & Raghav, 2026); both hold agent-generated experience in a central shared store, not in individuals' own memories. A SECI-based multi-agent framework converts the tacit elements of individuals' generative-AI use into explicit knowledge shared across an organisation (Hino et al., 2026), and SS-ZKR proposes, as a conceptual design without a prototype, privacy-preserving semantic routing of agent payloads across organisational trust boundaries (Touheed, 2026). Hivemind differs in exchanging knowledge held in individuals' own memories between their agents under bilateral trust; SS-ZKR addresses routing and might complement it.

Expert-finding systems are worth addressing directly, since they appear to solve a similar problem. Systems described by Yimam-Seid and Kobsa (2003) and Maybury (2006) automatically build expertise directories from digital traces - email metadata and document authorship - without requiring manual curation. This addresses the TMS maintenance problem through passive inference. These systems work from explicit artefacts, so they locate who is likely to know something; they do not capture or transfer the tacit knowledge itself, which still depends on a conversation between the two people. The difference with Hivemind is structural, not superficial. These systems require centralised access to workplace content, administrator-level access control, and institutionally mandated data collection. Hivemind infers expertise profiles from personal AI memory that is shared only peer-to-peer, under access scopes each person defines, and never with a central repository. The distinction is individual data sovereignty, which this paper hypothesises may influence whether people contribute in the first place (Sections 4.1.1 and 4.1.2).

## 3. Research Aims and Objectives

### *Aim*

To propose a federated protocol enabling AI agents to share tacit knowledge across team and departmental boundaries within an organisation, operationalising the Socialisation phase of the SECI model asynchronously and independently of human co-presence, and to show by a proof of concept that the protocol is technically realisable.

### *Objectives*

1. To identify and characterise the gap in current AI tooling with respect to tacit knowledge exchange using the SECI framework as a theoretical lens.
2. To evaluate the design space of centralised versus decentralised architectures for AI-mediated tacit expertise sharing.
3. To propose a federated protocol (Hivemind) addressing this gap, including trust establishment, semantic routing, cross-nomenclature bridging, and network-aware referral.
4. To show, through a proof-of-concept implementation, that the protocol can be implemented as specified.
5. To identify open research questions regarding benefit measurement and the limitations of the proposed approach.

### *Scope*

This paper proposes the system design and shows, by prototype, that it is technically realisable. Empirical measurement of corporate benefit is explicitly left to future work (Section 7.2). The formal treatment of data sovereignty across enterprise hierarchy levels is likewise reserved for a separate future research programme (Section 8.3).

## 4. Solution Proposal: Design Considerations

### *4.1 Design Space*

Before proposing a specific architecture, this paper evaluates the two principal design alternatives for an AI-mediated tacit knowledge sharing system.

#### *4.1.1 Centralised Hub Approach*

The choice between centralised and federated architectures is not universal; it depends on workplace context, regulatory environment, and the nature of the knowledge being managed. This paper presents both options with their evidenced advantages and limitations, recognising that different organisations will reach different conclusions based on their specific circumstances.

In a centralised approach, a single institutional server holds all employees' AI-captured knowledge. AI agents query the hub directly. This approach offers genuine advantages:

- It enables organisation-wide search with consistent access (Lee & Hong, 2002; Desouza et al., 2003).
- It simplifies governance, auditability, and administration.
- Critical knowledge remains accessible regardless of individual employee availability or working hours - a significant advantage for time-sensitive queries across teams and time zones.

The problem is that the same literature which motivates the tacit knowledge problem suggests that centralisation may make it worse. Reviews of knowledge-sharing barriers identify individual, organisational and technological factors that discourage contribution (Riege, 2005), and the effort of codifying knowledge for a central repository is a cost whose effect on contribution depends on trust (Kankanhalli et al., 2005). Monitoring of employees is associated with increased stress and reduced job satisfaction (Kalischko & Riedl, 2021), and in one study of military officers, supervisors' close monitoring had a negative impact on employees' knowledge sharing by undermining the leader-member relationship (Son et al., 2017). The evidence is mixed, however: Pei et al. (2026) found that monitoring intended to develop employees was positively associated with knowledge sharing, while monitoring intended to deter had no significant effect. This paper therefore proposes, as a hypothesis for future testing, the KM Surveillance Paradox: where a centralised system makes every contribution visible and searchable, that visibility may suppress the contribution behaviours that make the search worth doing.

#### *4.1.2 Decentralised Federated Approach*

In a decentralised approach, each employee maintains their own personal knowledge store. Knowledge never leaves the owner's system without explicit consent. Queries are routed peer-to-peer. This model offers:

- Employees show higher willingness to contribute when data sovereignty is preserved (Razmerita et al., 2009), consistent with social-capital explanations of why individuals contribute to knowledge communities at all (Wasko & Faraj, 2005).
- It offers greater resilience and scalability (Maier & Hädrich, 2006).
- It is expected to be cheaper to deploy and operate, because no central store has to ingest, index and secure every employee's memory and only a lightweight coordination server is needed; indexing compute is distributed across the nodes rather than eliminated, and this paper does not measure that cost.
- It is expected to reduce the visibility of individual contributions to a central party, visibility that Section 4.1.1 identifies as a possible deterrent to sharing.

- Networks form organically without central authority (Gupta & Bostrom, 2006).

The downsides are real. Discoverability is the hardest: without a central index, finding who has the knowledge you need is non-trivial (Ranjan et al., 2008; Maier & Hädrich, 2006). Inconsistent vocabulary across nodes makes retrieval even harder (Santoro et al., 2006). The network also starts empty - early users get little value until enough connections exist (Tsui, 2002). And when a peer's device is offline, their knowledge is simply not reachable.

*4.1.3 Rationale: The Hivemind Architecture*

The federated architecture proposed here (Figure 3; compared against the centralised alternative in Table 1) is a direct response to the decentralised model's known failure modes rather than a compromise between the two approaches. The coordination server is kept lightweight by design - it stores routing metadata only, never knowledge content - because, under the KM Surveillance Paradox hypothesis (Section 4.1.1), central visibility of contributions is the likely deterrent. Discovery complexity is addressed through LLM-generated alias bridging. Cold-start is mitigated, once at least one trust edge exists, through network-aware referral (Section 5.5; the residual limitation is discussed in Section 8.4). Ontological inconsistency is addressed through synonym clusters. The decentralised model's core advantage identified in Section 4.1.2 - higher contribution under data sovereignty - is preserved.

Organisations seeking to improve tacit knowledge transfer through centralisation may inadvertently deepen the problem they seek to solve, as hypothesised under the KM Surveillance Paradox (Section 4.1.1); Table 2 maps the documented barriers from Sections 1, 2, 4.1.1 and 8.6 to the specific Hivemind mechanisms intended to address them.

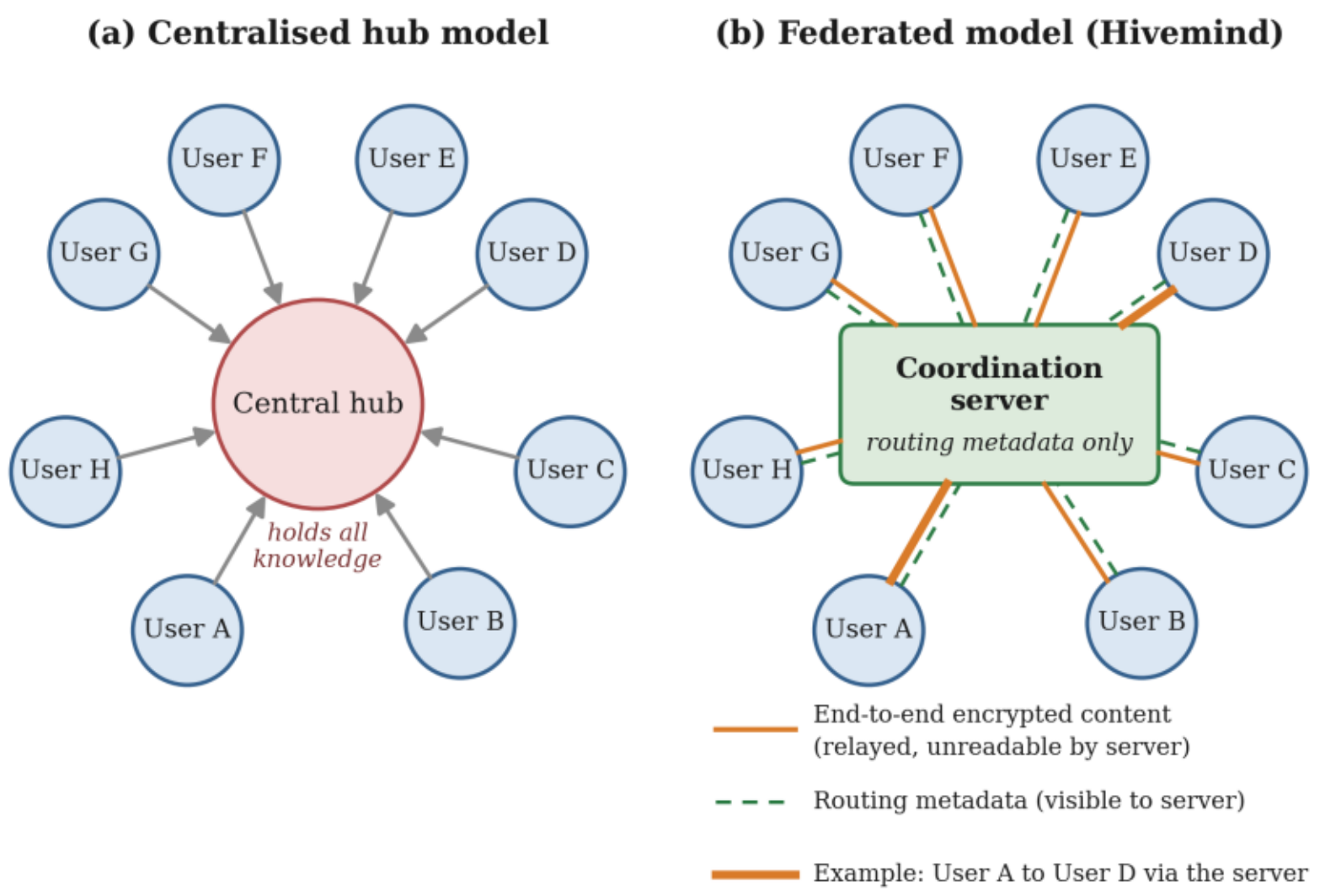

*Figure 3: Centralised vs. federated architecture. In the centralised model (left) a single server holds all knowledge; in the federated model (right) a lightweight coordination server stores only routing metadata.*

*Table 1: Architecture comparison - centralised versus federated properties*

| Property | Centralised Hub | Federated (Hivemind) |
|---|---|---|
| Knowledge location | Central server | Each user's own system |
| Query routing | Direct hub query | Peer-to-peer via routing cache |
| Privacy | Hub operator sees all | End-to-end encrypted; hub sees only metadata |
| Availability | Always-on regardless of peers | Requires peer to be online for content |
| Employee willingness | May be suppressed by surveillance | Expected to be higher due to sovereignty |
| Cold start | Immediate (pre-populated) | Requires network formation |
| Maintenance | Requires curation effort | Automatic via periodic sync |
| Deployment and operating cost | Large central store to host, secure and administer | Light coordination server; each user runs a client |
| Compute and token cost | Central indexing of all memory, paid once | Spread across nodes; repeated per peer pair |

### *4.2 Data Sovereignty and Privacy Considerations*

The coordination server observes three categories of information: traffic metadata (which nodes contacted which peers and at what times), synonym cluster declarations contributed by users (revealing the semantic vocabulary of organisational projects and their relationships), and the names of the knowledge domains that each pair of peers agrees to share at handshake time, recorded as part of the trust graph. It also holds public keys and a token revocation list, so it can see when a trust relationship is created or revoked. It does not observe the content of any query, the text of any memory record, or any individual record of reasoning or decision, in keeping with the principle that federated systems should disclose no more than a given request strictly requires (Khanzadeh et al., 2026). Knowledge stays in each user's own local memory store, which this design does not itself encrypt, and is sent only to authenticated peers holding valid capability tokens, encrypted end to end in transit.

Three threat properties are worth stating plainly.

Honest-but-curious coordination server. The coordination server relays encrypted traffic and could be operated by a party curious about its users. It cannot read query text, responses or memory content, because these are end-to-end encrypted between peers; *what it can learn is limited to the metadata listed above*. Organisations that regard this residual metadata as sensitive should host the server themselves.

Compromised or over-curious peer. A peer can read only the knowledge domains named in the capability token it holds, and private content is excluded before it leaves the owner's node (domains the owner marks as private are never exposed, and content tagged private is stripped from responses). A peer cannot, however, be prevented from copying or onward-sharing what it has legitimately received: Hivemind *limits what is exposed, not what a trusted recipient does with it*.

Stolen tokens. Capability tokens are signed, scoped to named domains and carry a unique identifier, so the owner can revoke one immediately and the peer's client rejects further requests within seconds (Section 5.2). A token stolen before revocation grants only the access it was scoped to; this bounds the damage but does not prevent it, which is why revocation practice is part of the ongoing maintenance described in Section 9.

In regulated industries, mandatory audit trail requirements may conflict with this architecture; this constraint is discussed in Section 8.7.

### *4.3 Knowledge Quality and Provenance*

What the system returns is only as good as what was captured. Topic aliases are inferred from the content of AI memory, so gaps and errors in the underlying record carry through. Attribution shows where the knowledge came from, but not whether it is correct - verifying that is the querier's job. This mirrors how TMS works in practice: the directory tells you who to ask; it does not guarantee the answer.

Where accuracy verification is required, retrieved knowledge can be cross-validated against externalised knowledge sources - corporate KMS tools, web search, or academic repositories - accessible via the same integration layer (MCP). This hybrid approach enables AI agents to triangulate tacit peer knowledge against explicit institutional knowledge, surfacing conflicts or confirmations as part of the response.

*Table 2: Documented Barriers Mapped to Hivemind Mechanisms*

| Documented Barrier | Source(s) | Hivemind Mechanism | Section |
|---|---|---|---|
| Possible surveillance-driven contribution suppression (hypothesis) | Riege, 2005; Kankanhalli et al., 2005; Son et al., 2017 | Federated sovereignty - content stays with its owner unless shared under a capability token | 4.1.3 |
| Vocabulary heterogeneity across teams | Edgington et al., 2004; Blomqvist & Öhgren, 2008 | LLM alias bridging (structural + semantic) | 5.4 |
| TMS maintenance failure | Peltokorpi, 2008 | Automatic routing cache via daily synchronisation | 5.3 |
| Search-transfer gap | Hansen, 1999 | Cache-assisted referral routing | 5.5 |
| Co-presence requirement for tacit exchange | Nonaka & Von Krogh, 2009; Williams, 2012 | Asynchronous peer query via persistent relay | 5.1/5.6 |
| Causal ambiguity in knowledge identification | Szulanski, 1996 | Acknowledged limitation - not addressed by current design | 8.6 |

## 5. Proposed System: Hivemind

### *5.1 Architecture Overview*

Hivemind follows a federated hub-and-spoke topology (Figure 4). A lightweight coordination server acts as a relay and trust broker. Personal AI memory systems are the spokes. The server stores only: a trust graph (including the domains shared on each trust edge), public keys, synonym clusters, and a token revocation list - never knowledge content. Clients connect outbound to the server via persistent WebSocket; the server never initiates connections to clients, enabling operation behind corporate Network Address Translation (NAT) and virtual private network (VPN) set-ups without port exposure. The mapping of these mechanisms to the SECI model is shown in Figure 5 and discussed in Section 7.1.

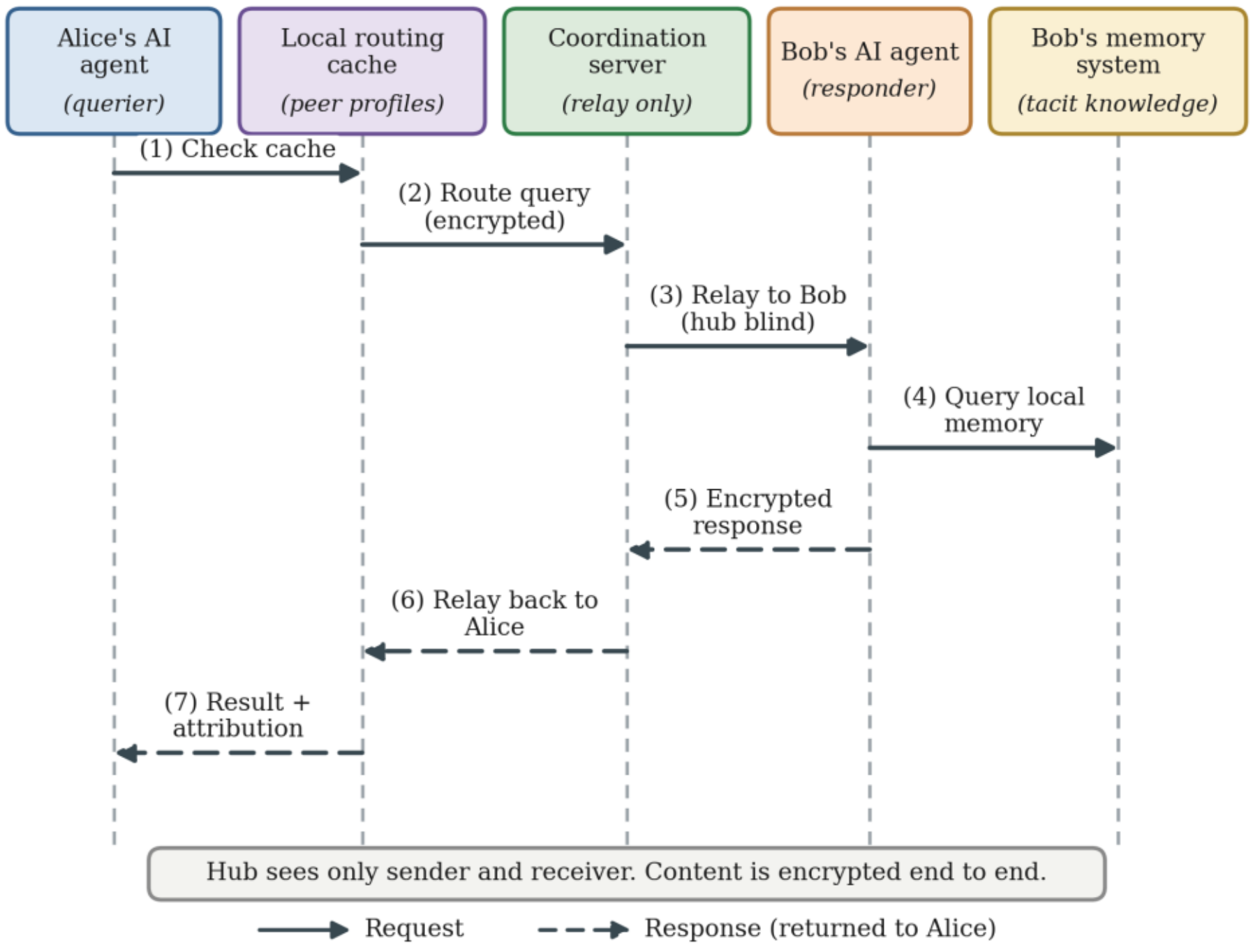

*Figure 4: Hivemind query flow. A seven-step sequence showing how a query moves from Alice's AI agent through the coordination server to Bob's memory system and back, with the routing cache maintained on both sides.*

### 5.2 Trust Establishment

Trust is established through a one-time bilateral handshake at knowledge-domain granularity: each party independently selects which knowledge domains to share before any content is exchanged. The coordination server issues scoped JWT (JSON Web Token) capability tokens specifying: subject (token holder), target (whose memory is accessible), permitted domains, and token type. Tokens are long-lived by default and immediately revocable. Revocation blacklists the token ID; the peer's client rejects all further requests within seconds.

### 5.3 Semantic Indexing and Periodic Synchronisation

Each client maintains a two-tier per-peer routing cache. The first tier is populated by periodic structural synchronisation (daily): domain names and LLM-generated aliases (4-5 plain-language descriptors per domain, derived from structure only, never content) are exchanged with each trusted peer via the encrypted relay. This mirrors time-to-live (TTL) caching in the Domain Name System (DNS): knowledge domain coverage changes slowly; daily refresh is proportionate to the rate of change and minimises privacy exposure. The second tier is populated by deep semantic indexing at handshake time (one-time): shared domain content is sent, encrypted, to the requesting peer, whose client uses an LLM to extract 8-12 semantic topic descriptors, which are cached as deep aliases and shared back to the peer.

### *5.4 Cross-Nomenclature Alias Bridging*

Motivated by the vocabulary heterogeneity documented in the KM literature (Edgington et al., 2004; Blomqvist & Öhgren, 2008), the alias system operates at both structural and semantic levels. Users may additionally declare synonym clusters ('compound registry = molecule catalogue = chemical inventory'), which the coordination server stores as equivalence classes benefiting all network members - a lightweight implementation of ontology alignment approaches proposed in the literature (Hameed et al., 2004).

### *5.5 Network-Aware Referral Routing (AI-Layer TMS)*

When a peer's routing cache lacks coverage for a query, the peer may return a referral pointer to a second-degree connection whose profile suggests relevance. This computationally implements the directory component of Transactive Memory Systems (Wegner, 1987) - the shared awareness of who knows what - with automatic maintenance that directly addresses the effort-based failure mode established by Peltokorpi (2008). The allocation component (routing new knowledge to appropriate holders) and calibration component (expertise credibility signalling) remain outside current scope, a direction for future work consistent with Ren and Argote (2011).

### *5.6 Protocol Message Specification*

Three message types govern Hivemind peer communication; each QUERY is answered by a RESPONSE message that carries the encrypted result. The handshake initiation establishes a trust relationship between peers:

```
{
 "type": "HANDSHAKE_REQUEST",
 "from": "alice",
 "to": "bob",
 "proposed_domains": ["engineering", "auth-systems"],
 "public_key": "base64..."
}
```

Domain synchronisation transmits alias mappings (structural_aliases, and the LLM-generated aliases carried in lm_aliases) between consenting peers on the daily schedule described in Section 5.3 (the example is abbreviated):

```
{
 "type": "SYNC_PAYLOAD",
 "from": "alice",
 "to": "bob",
 "domains": [
 {
  "name": "compound-registry",
  "structural_aliases": [
   "registry service", "compound database", "backend"
  ],
  "lm_aliases": [
   "molecule catalogue", "chemical inventory service"
  ]
 }
 ],
 "encrypted": true
}
```

Query messages carry the embedding of the question and a capability token authorising access:

```
{
 "type": "QUERY",
 "from": "alice",
 "to": "bob",
 "question_embedding": "[vector]",
 "capability_token": "JWT..."
}
```

Capability tokens are signed JWTs carrying the following claims: sub (the token holder), target (whose memory is accessible), permitted_domains (list of accessible domain names), token_type (either permanent or temporary), and jti (a unique identifier supporting revocation).

The query flow proceeds as follows: Alice submits a QUERY to the coordination server, which relays it to Bob’s client. Bob’s client decrypts the message, queries its local memory store, and returns a RESPONSE to the coordination server, which relays the encrypted result back to Alice for decryption and surfacing. No plaintext memory content transits the relay at any stage.

Computational complexity: query routing is O(n) in the number of trusted peers at query time; daily domain synchronisation is O(k) in the number of domain aliases per peer.

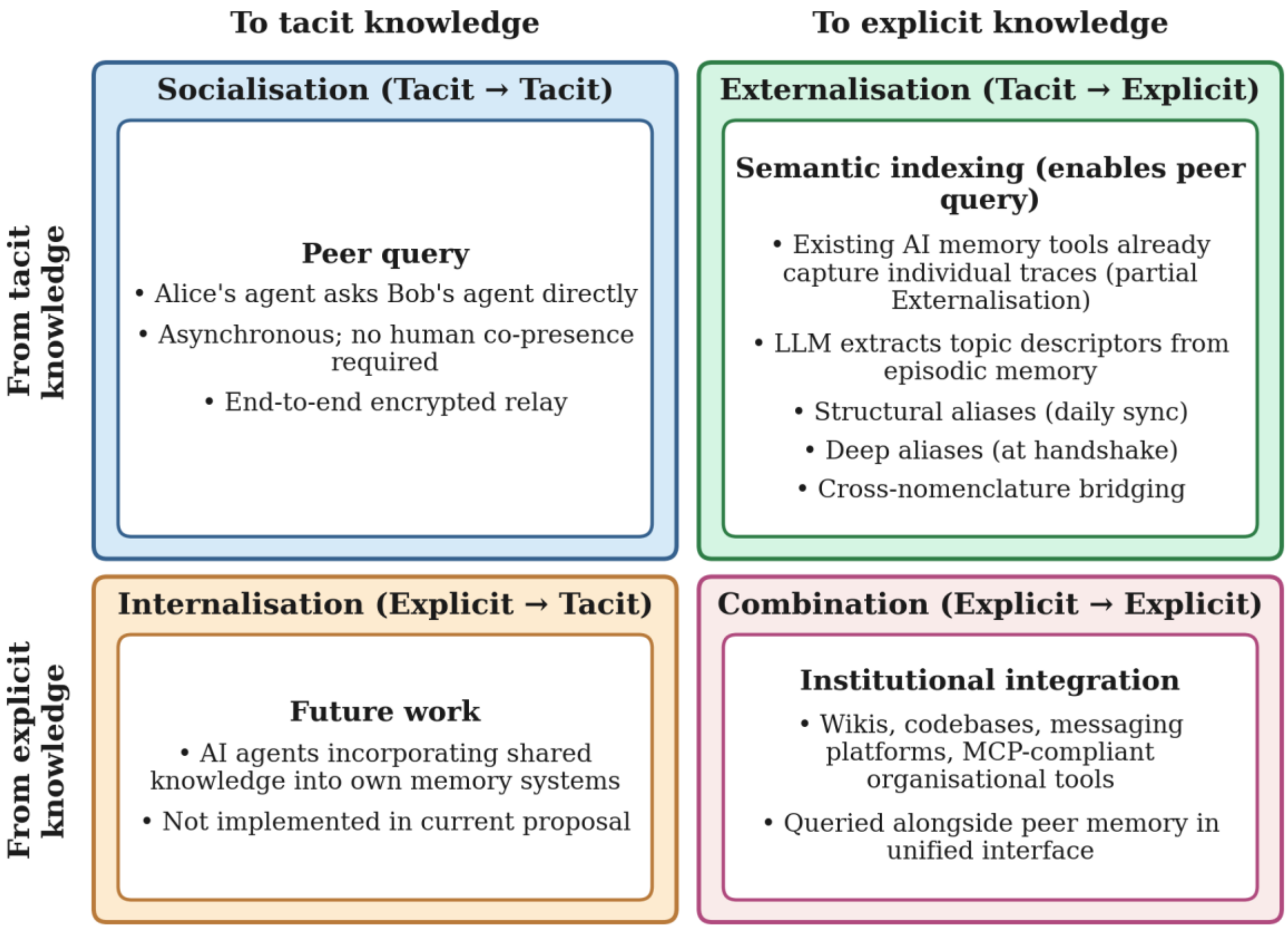

*Figure 5: Mapping the SECI model to Hivemind mechanisms. Combination and individual-level Externalisation are already served by existing AI tooling; Socialisation is addressed by peer query, with semantic indexing as its enabling mechanism; Internalisation of shared knowledge remains future work.*

## 6. Proof of Concept

The proof of concept presented in this section constitutes a Design Science Research (DSR) artefact demonstration in the sense of Hevner et al. (2004) and Peffers et al. (2007). This paper follows the DSR research methodology: problem relevance and knowledge gap are established in Sections 1-2; the design artefact - the Hivemind protocol and federated architecture - is specified in Sections 4-5; artefact demonstration via a proof-of-concept implementation is presented here; and research contribution is assessed in Section 7. Consistent with DSR evaluation norms (Hevner et al., 2004), this stage constitutes artefact demonstration rather than empirical validation; structured expert evaluation of the deployed artefact is the intended empirical evaluation step, reserved for future work.

### *6.1 Implementation and Test Scenarios*

The purpose of the proof-of-concept implementation is to establish that the proposed protocol is technically realisable, not to measure enterprise benefit. The implementation comprises four components: a coordination server (FastAPI, SQLite, WebSocket relay, Docker image; 39 unit tests passing), a client MCP server (Python, end-to-end encryption via PyNaCl, six tools exposed to AI assistants; 45 unit tests passing), a web visualisation layer (D3.js trust graph and knowledge map with knowledge concentration risk display), and a Docker Compose single-machine test harness seeding three memory instances - Alice, Bob, and Carol - with deliberately mismatched domain naming to validate cross-nomenclature routing.

Nine end-to-end scenarios (Table 3) were executed against the test harness. All nine passed; these are functional checks on authored test data, with no real users and no baseline, and they show that the design works as specified, not that it works in practice.

*Table 3: End-to-end test scenarios and outcomes*

| **Scenario** | **Nodes Involved** | **Test Purpose** | **Outcome** |
|---|---|---|---|
| Trust establishment | Alice ↔ Bob | Verifies bilateral handshake and token issuance | PASS |
| Periodic synchronisation | Alice, Bob | Verifies structural alias exchange completes within 2 s | PASS |
| Semantic indexing | Alice indexes Bob | Verifies deep alias extraction and sharing back to peer | PASS |
| Query routing | Alice queries "compound registry authentication" | Verifies result returned from Bob's memory with source attribution | PASS |
| Cross-nomenclature resolution | Alice queries "pharmaceutical compound registry kotlin" | Verifies Bob's "drug-discovery" domain matched via deep semantic alias despite zero lexical overlap | PASS |
| Private domain exclusion | Alice queries Carol's private domain | Verifies zero results returned regardless of content relevance | PASS |
| Private-marker stripping | Alice queries domain containing private markers | Verifies marked content is absent from the response | PASS |
| Token revocation | Bob revokes Alice | Verifies Alice's subsequent query returns no result from Bob within seconds of revocation | PASS |
| Referral chain routing | Alice queries Carol-only knowledge | Verifies Bob's referral pointer successfully routes Alice to Carol | PASS |

The cross-nomenclature scenario warrants specific description, as it directly tests the mechanism motivated by the vocabulary heterogeneity literature (Section 2.2). Alice's memory instance contains a domain named "compound-registry"; Bob's memory instance contains a domain named "drug-discovery". Both domains describe the same engineering system using distinct team vocabularies with zero lexical overlap in domain names. A query of "pharmaceutical compound registry kotlin" issued from Alice's agent returned a correctly attributed result from Bob's "drug-discovery" domain, routed via the deep semantic aliases generated during the one-time indexing phase. No synonym cluster declaration was required; semantic similarity alone was sufficient to bridge the nomenclature gap.

All nine scenarios were executed within a single-machine Docker Compose environment. A single local measurement put content relay round-trip latency at 50-120 ms, which says nothing about production latency but is consistent with the assumption that knowledge queries are asynchronous background operations rather than synchronous user-facing calls.

The paper makes no claim about workplace benefit from this implementation. The proof of concept demonstrates that the protocol functions as designed across trust, routing, privacy, revocation, and referral dimensions. Benefit measurement is addressed in Section 7.2.

### *6.2 Proposed Expert Evaluation*

Following Design Science Research conventions (Hevner et al., 2004), a structured expert evaluation is proposed as future empirical work to validate the Hivemind artefact beyond the technical feasibility demonstration presented in Section 6. The evaluation will recruit 5-8 Knowledge Management practitioners from IT organisations, selected to represent variation in institutional size and knowledge-sharing practices. Data collection will employ semi-structured interviews guided by three evaluative dimensions: utility (does the design address a real and significant problem in distributed knowledge work?), completeness (does the artefact cover the key requirements identified in the literature review?), and feasibility (could this architecture realistically be deployed in an organisational context?). Assessment dimensions are grounded in the Technology Acceptance Model (Davis, 1989), specifically the perceived usefulness and perceived ease of use constructs. The current artefact demonstration establishes technical feasibility as a prerequisite for, and not a substitute for, this prospective evaluation.

## 7. Discussion

### *7.1 Hivemind as Digital Ba: SECI Mapping*

Peer query in Hivemind most closely resembles Nonaka and Takeuchi's (1995) Socialisation quadrant (Figure 5): tacit-knowledge residues embedded in one individual's practice reach another without passing through explicit codification. However, this mapping must be held loosely. Socialisation, as Nonaka and Von Krogh (2009) clarify, is grounded in embodied co-presence - apprenticeship, shared physical activity, observation, and imitation - through which practitioners absorb know-how that resists articulation. Hivemind produces no such embodied exchange. What Hivemind does instead is operationalise the corporate function historically served by Socialisation: enabling tacit-knowledge residues embedded in one individual's work context to reach another individual across distance and time, independent of physical co-presence (Tsoukas, 2003). This is a functional approximation, not a literal replication of Nonaka's quadrant - the mechanism and phenomenology differ fundamentally, but the enterprise problem addressed is the same: bridging the gap between contextual knowledge held by one practitioner and the awareness of a colleague who would benefit from it.

This distinction between functional approximation and literal replication matters because the conditions Nonaka requires for Socialisation are rarely met in real organisations. Nonaka et al. (2000) list the enabling conditions: mutual trust, active empathy, lenience in judgement, courage, and access to help. Research has consistently shown how far these conditions fall short in practice - particularly in distributed, high-workload environments (Riege, 2005; Foos et al., 2006; Scarso & Bolisani, 2008). AI agents are not subject to these conditions at all. There is no ego, no status anxiety, no reluctance to approach a senior colleague's knowledge store. An AI agent treats another agent's episodic memory as a data source. It does not hesitate, does not feel intimidated, and has no incentive to withhold. The obstacles that have historically blocked Socialisation - hierarchy, busyness, reticence - do not apply. What replaces them are technical and governance questions: How do agents find relevant peers? Who controls what gets shared? How reliable is AI-extracted knowledge? These questions are practical rather than interpersonal, which suggests the approach may be tractable where the human-centric approach has struggled for decades.

This framing sits in tension with the enabling conditions argument advanced by Nonaka and Von Krogh (2009), who maintain that technical infrastructure cannot substitute for the relational preconditions - mutual trust, active empathy, and tolerant judgement - without which tacit knowledge will not be offered or safely received. Hivemind's design acknowledges this constraint rather than contesting it: bilateral peer connection requires explicit consent from both individuals, meaning the protocol presupposes a pre-existing relational foundation rather than generating one, and reduces the friction of knowledge access only where that foundation already holds.

The critical contribution relative to He et al.'s (2026) conceptual framework, and to federated agent-knowledge approaches more broadly (Hou, 2026; Li et al., 2026), is that Hivemind creates a federated digital Ba - not a single shared space, but a network of personal spaces interconnected by trust edges, spanning multiple individuals and their AI agents simultaneously. Knowledge remains sovereign and is accessed on demand. This is architecturally essential for the same reason given in Section 4.1.2: data sovereignty is expected to sustain contribution.

### *7.2 Measuring Benefit: Open Research Questions*

The paper does not measure workplace benefit. Constructing a valid measurement methodology for knowledge-sharing improvement involves substantial methodological challenges that constitute a research programme in their own right.

Qualitative approaches include: structured interviews on perceived usefulness (Technology Acceptance Model; Davis, 1989), knowledge-seeker satisfaction, and expert assessments of retrieval quality. Quantitative approaches include: knowledge retrieval accuracy against ground-truth query sets; time-to-competence for new team members; reduction in parallel knowledge-rediscovery events; and reduction in knowledge concentration risk over time. Counterfactual construction is difficult; knowledge quality is hard to operationalise; user adoption creates confounds. These questions are explicitly left open for future work, and this paper encourages empirical follow-on studies.

## 8. Limitations

### *8.1 Intentional Knowledge Withholding*

Hivemind facilitates sharing of knowledge; it does not compel it. Deliberate withholding of tacit knowledge (Connelly et al., 2012; Riege, 2005) will not be overcome by any technical system. The federated, sovereignty-preserving design is intended to reduce the surveillance concern that may amplify knowledge hiding in centralised systems, but intentional withholding at the individual level remains outside the scope of technical intervention. Social, managerial, and cultural interventions remain necessary complements.

### *8.2 Semantic Aliasing Boundary (Future Work)*

The cross-nomenclature alias system bridges name variation within the same concept domain. It does not address the case where different teams develop architecturally similar projects that share surface terminology but represent distinct knowledge - for example, two separate front-end/back-end systems in unrelated domains that share architectural vocabulary. Identifying conceptual similarity across genuinely different domains is a substantially harder problem requiring topic modelling, knowledge graph alignment, or domain expert input. This is left as a separate contribution for future research.

### *8.3 Data Sovereignty Hierarchy (Future Work)*

A complete treatment of data sovereignty in institutional knowledge exchange must address multiple layers: organisation-wide public knowledge, department-specific knowledge, team-specific knowledge, and individual private knowledge. Each layer requires different access control models, trust calibration mechanisms, and governance frameworks. The current proposal treats this at a binary level (shared/not shared within a handshake scope). A rigorous multi-layer sovereignty framework lies outside the scope of this paper.

### *8.4 Cold Start (Empty Network)*

Hivemind is only useful once you are connected to colleagues, and every connection is a handshake that two people agree by hand. A new user therefore starts with an empty network and gets little value until a few trusted colleagues have connected. The coordination server keeps no list of people or topics to browse, so a stranger cannot be found and nobody can be approached without an existing relationship. This is deliberate, because it keeps trust in human hands and keeps the server blind to content. Referral (Section 5.5) helps once a user is connected, because a colleague can point to a colleague of theirs, and Section 10 advises starting with two or three trusted colleagues. Users without an established professional network still gain little at first, and this paper offers no remedy beyond the opt-in public topic index proposed as future work.

### *8.5 Offline Peers and Content Freshness*

When a peer's client is disconnected, their knowledge domains are unavailable for content queries (though cached routing metadata remains usable for routing decisions). Unlike a centralised model, peer unavailability directly blocks content retrieval. The daily synchronisation window means the routing cache may lag new knowledge domains by up to 24 hours. Indexing compute and assistant tokens are borne by every node and repeated for each trusted pair, so total compute may not fall; this was not measured.

### *8.6 Causal Ambiguity*

Szulanski (1996) identified causal ambiguity - the inability of either the source or the recipient to fully articulate why knowledge works and what conditions enable its transfer - as one of the major barriers to knowledge transfer within organisations, alongside the recipient's lack of absorptive capacity and an arduous relationship between source and recipient, and found these knowledge-related barriers to outweigh motivational ones. Hivemind reduces access friction for knowledge that can be accurately described, semantically routed, and recognised as relevant by at least one party at query time; it does not resolve causal ambiguity for knowledge whose relevance neither the source agent nor the querying agent can recognise, a category that, by Szulanski's (1996) analysis, accounts for a substantial share of knowledge-transfer failures regardless of the quality of the channel.

### *8.7 Audit Trail and Regulated-Industry Constraints*

In regulated industries, a cryptographic blindness design is incompatible with mandatory audit trail requirements. Hivemind's data sovereignty architecture is appropriate for voluntary expertise sharing networks; organisations under audit obligations would require administrator-accessible audit logs before deployment. The local memory store is not encrypted by the system; protecting it is left to the host (for example disk encryption).

It is worth noting that the federated model is not the universal answer. Organisations operating around the clock across time zones, or those in highly regulated industries requiring comprehensive audit trails, may find that a centralised architecture better serves their operational requirements, notwithstanding the possible contribution-suppression effects discussed in Section 4.1.1. The design considerations presented in Section 4 are intended to support, not replace, organisational judgement about the appropriate architecture for a given context.

## 9. Corporate Deployment Considerations

Deploying Hivemind at enterprise scale requires three distinct roles. The Coordination Server Administrator is responsible for provisioning and maintaining the hub infrastructure, managing TLS certificates, monitoring availability, and applying software updates; this role does not require access to knowledge content and can be fulfilled by an existing platform or DevOps team. The Trust Graph Steward oversees the organisation's trust graph structure: they may establish handshakes on behalf of teams as a convenience, communicate norms around which knowledge domains should be shared across team boundaries, and curate the central synonym cluster vocabulary that determines cross-team query reach. In smaller organisations this role may overlap with the administrator; in larger deployments it benefits from being a designated function within a team knowledge management mandate. Individual Node Users are responsible for installing and configuring their own clients, deciding which knowledge domains to expose within each handshake, and keeping their local AI memory system populated with AI-assisted work.

Ongoing maintenance requires three recurring activities: periodic review of synonym cluster accuracy, as project terminology drifts over time; renewal or explicit revocation of trust relationships as team structures change through reorganisations and staff transitions; and manual update of domain structure when domains are renamed or reorganised, since the alias refresh mechanism handles minor nomenclature drift automatically but cannot infer structural intent.

When an employee leaves the organisation, their node transitions to a ghost archive status: it becomes read-only for a configurable retention period, preserving accumulated knowledge for authorised peers while accepting no new writes. A future offboarding protocol, identified for further investigation, would automate the identification of orphaned knowledge held in ghost archives and prompt designated successors to absorb it before the retention window expires.

## 10. Implications for Practitioners

Team leads deploying Hivemind for the first time should begin with two or three bilateral handshakes among trusted colleagues on a specific, bounded project rather than attempting an organisation-wide rollout. The first month should be treated as alias calibration rather than a production search phase: the primary value at this stage lies in discovering what terminology each colleague uses for shared concepts, not yet in retrieving answers from the network.

Hivemind presupposes basic interpersonal trust - it does not create it. The protocol enables knowledge sharing; it cannot supply the social willingness to share. In organisations where knowledge hoarding is incentivised, or where strong information-silo norms prevail, contribution rates are likely to be low regardless of the technical infrastructure. Adoption should therefore be preceded by an honest assessment of whether the team's collaborative culture supports the model.

Practitioners should be equally clear about what Hivemind cannot do. It does not replace documentation culture: it augments it, lowering the cost of capturing tacit knowledge that would otherwise never be written down. It cannot surface knowledge that was never externalised through an AI-assisted session - memory that remains entirely inside a person's head is invisible to the network.

This paper does not measure return on investment, but the value of the network is expected to grow with its size. A first handshake connection is likely to yield modest incremental value. As connections accumulate, the knowledge topology may become qualitatively different from any single node: rare expertise could become discoverable on demand, single-point-of-failure knowledge concentrations could become visible before they become crises, and the organisation's collective reasoning history could begin to constitute a navigable institutional resource.

## 11. Future Work

- Cold-start discovery via opt-in public topic index.
- Direct peer connections removing the coordination server as relay.
- Cross-organisation federation between independently operated instances.
- Formal privacy analysis, including zero-knowledge approaches to semantic routing.
- Longitudinal empirical evaluation of workplace knowledge-sharing improvement.
- Multi-layer data sovereignty framework (Section 8.3).
- Internalisation mechanisms: AI agents updating their own memory from shared knowledge.
- Cross-domain disambiguation of surface-similar terminology (Section 8.2).

## 12. Conclusion

The problem of tacit knowledge inaccessibility in organisations, and of deliberate hoarding in particular, is neither new nor confined to the moment of employee departure. It is an ongoing failure that manifests daily in knowledge silos, parallel rediscovery, extended onboarding, and fragile dependence on key individuals. Despite decades of research into knowledge management systems and transactive memory, and despite significant recent advances in AI-powered access to explicit institutional knowledge, the Socialisation phase of Nonaka and Takeuchi's SECI model has remained largely intractable in distributed digital organisations - precisely because it historically required co-presence, synchronous interaction, and deliberate human effort.

The paper has proposed Hivemind: a federated protocol in which AI agents participate in a digital Ba on behalf of their users, enabling tacit knowledge exchange asynchronously, across team and departmental boundaries, and independently of human co-presence. The approach is grounded in thirty years of knowledge management theory and directly addresses the failure modes of manually maintained Transactive Memory Systems (which fall into disuse under workload pressure) and the hypothesised failure mode of centralised knowledge repositories (which may suppress contribution through surveillance).

The proposed federated architecture offers a functional resolution, in the sense established in Section 7.1, to the hypothesised central paradox of centralised KM design: the very monitoring that enables search may also suppress the contribution that makes search valuable. By preserving data sovereignty and routing knowledge access through bilateral trust rather than central administration, Hivemind creates conditions in which voluntary knowledge transfer is more likely to occur - while simultaneously operationalising TMS semantics at the AI agent layer through an automatically maintained peer routing cache.

The cross-nomenclature alias bridging mechanism addresses a persistent, well-documented barrier to inter-team knowledge transfer: vocabulary heterogeneity between communities of practice. The network-aware referral mechanism extends knowledge discovery beyond direct connections, operationalising Wegner's transactive memory concept computationally and without the maintenance burden that causes human TMS implementations to fail.

A proof-of-concept implementation shows that the protocol can be implemented as designed. The paper explicitly does not claim to measure corporate benefit - that measurement programme remains future work, with both qualitative and quantitative methodological approaches identified.

The paper recognises the limits of technical solutions to social and enterprise problems. Hivemind cannot compel an employee who chooses to withhold knowledge, resolve the semantic concept drift that occurs when similar-sounding projects represent genuinely different domains, or substitute for the trust and cultural conditions that motivate sharing of knowledge in the first place. These are acknowledged limitations, and each points to a distinct research programme.

What Hivemind does offer is a reduction in the structural friction that prevents willing knowledge sharers from sharing. When episodic AI memory captures residues of what individuals know, Hivemind creates the shared digital Ba through which that knowledge can flow - without the synchronous co-presence that has always been Socialisation's limiting constraint. In doing so, it proposes a technically feasible path towards the vision He et al. (2026) described conceptually: AI as a participant in workplace knowledge creation - extended here from the individual human-AI dyad to the collective, federated, multi-agent scale at which institutional knowledge is, in practice, distributed.